\documentclass[sigconf]{acmart}
\usepackage{amsmath,amssymb,amsfonts}
\usepackage{algorithmic}
\usepackage{graphicx}
\usepackage{textcomp}
\usepackage{xcolor}
\usepackage{subfigure}
\usepackage{color,xcolor,soul,colortbl}
\usepackage{caption}

\AtBeginDocument{%
  \providecommand\BibTeX{{%
    \normalfont B\kern-0.5em{\scshape i\kern-0.25em b}\kern-0.8em\TeX}}}

\begin{document}

\title[A Multisensory Learning Architecture]{A Multisensory Learning Architecture for Rotation-invariant Object Recognition}


\author{Murat Kirtay}
\email{murat.kirtay@hu-berlin.de}
\affiliation{%
\institution{Adaptive Systems Group, Department of Computer Science, Humboldt-Universit{\"a}t zu Berlin}
\city{Berlin}
\country{Germany}  
}
\author{Guido Schillaci}
\email{guido.schillaci@santannapisa.it}
\affiliation{%
  \institution{The BioRobotics Institute, Scuola Superiore Sant'Anna}
  \city{Pontedera (PI)}
  \country{Italy}}

\author{Verena V. Hafner}
\email{hafner@informatik.hu-berlin.de}
\affiliation{%
  \institution{Adaptive Systems Group, Department of Computer Science, Humboldt-Universit{\"a}t zu Berlin}
  \city{Berlin}
 \country{Germany}
}


\begin{abstract}

This study presents a multisensory machine learning architecture for object recognition by employing a  novel dataset that was constructed with the iCub robot, which is equipped with three cameras and a depth sensor. The proposed architecture combines convolutional neural networks to form representations (i.e., features) for grayscaled color images and a multi-layer perceptron algorithm to process depth data. To this end, we aimed to learn joint representations of different modalities (e.g., color and depth) and employ them for recognizing objects. 
We evaluate the performance of the proposed architecture by benchmarking the results obtained with the models trained separately with the input of different sensors and a state-of-the-art data fusion technique, namely decision level fusion.  The results show that our architecture improves the recognition accuracy compared with the models that use inputs from a single modality and decision level multimodal fusion method.

\end{abstract}




\maketitle

\section{Introduction}
Vision-based robotics tasks (e.g., object recognition, vision-guided manipulation) rely on extracting useful features -- by processing different sensory readings--  of the perceived object and robustly combining them to be inputs of a machine learning  algorithm~\cite{eitel2015deepmm},~\cite{ruiz2018survey}.  In this study, we address the single-axis rotation invariant object recognition task by employing a multimodal machine learning architecture on a novel multimodal dataset, which was constructed with the iCub robot by using three cameras to obtain color images and a depth sensor. The proposed architecture -- i.e., intermediate representation fusion-- leverages convolutional neural networks (CNNs) and multilayer perceptron (MLP)  to process grayscaled color and depth inputs, respectively.  To test the recognition performance of the architecture, we benchmarked the results with different models that were trained by using color and depth data.  Additionally, we adopt a state-of-the-art fusion technique to benchmark the obtained result in multimodal settings.  

To further test the performance of our architecture and decision level fusion technique, we replaced one of the camera readings with randomly constructed data with the same size of images. Since hardware defects (e.g., miscalibrated sensors and unexpected noise sources) are common for robot experiments, we aim to assess our architecture in a setting that an employed sensor provides noisy readings for object recognition. 

In this work, we adopt the term modality as an employed sensory data that are associated with different aspects (e.g., the color and depth information) of the observed  phenomena~\cite{lahat2015multimodal},~\cite{kirtay2018multimodal}. In our setting, we used preprocessed color images and depth data as two different modality data -- though in biological agents, color and depth information emanates from the same sensory organ (i.e., eye) -- to perform object recognition.

A broad range of multimodal (i.e., multisensory) learning studies was reported in the fields of machine learning and robotics. Here, we merely introduce representative studies; however, we point out that more detailed information can be found in~\cite{kirtay2018multimodal},~\cite{Ramachandram2017survey},~\cite{Gao2019survey}.  Since the employed dataset is recently constructed, and there are no benchmarked studies yet, we present the literature studies in terms of the performed methods and fusion techniques.  To be concrete, we leverage the intermediate representations (i.e., features) of color information provided by three cameras to construct in-class shared representations.  We used the term in-class to underline the presentations for images, which were obtained by employing different cameras.  We then fused these representations with depth information processed by the multi-layer perceptron.

The authors of the study in~\cite{eitel2015multimodal} -- and similarly in~\cite{socher2012convolutional}-- implemented an architecture based on a convolutional neural network (CNN) to process depth and color modalities.  Two streams of CNN for depth and preprocessed color
data were merged in a fully connected layer before performing object recognition.  By doing so, the authors reported that the recognition performance was improved by applying a late fusion technique.  
In our architecture, we employed CNNs to process grayscaled color images that were obtained from different cameras. Since the employed dataset is modality-wise imbalanced --that is, the number of images is higher than corresponding depth data-- we first aimed at extracting the shared representations of the color modality.  Then, we combined the shared representations constructed by CNNs with the last layer of MLP activations by using depth data to build a joint representation of different sensory data.  Unlike the study in~\cite{eitel2015multimodal}, we did not preprocess depth data (i.e., colorization of the depth); instead, we test our architecture in a setting where using depth only modality is not performing well in object recognition.

In~\cite{wang2015mmss}, the authors first separately trained two deep CNN streams by using color and depth data. Then, the activations of the fully connected layers were combined with learning shared multimodal representations. The study shows that combining depth and color information via CNNs enhances object recognition performance.
The authors in~\cite{oh2017object} applied the decision level fusion technique by training different CNNs in a multisensor setting (i.e., lidar for distance and charge-coupled device for color images) for detecting and classifying objects.  The authors build a unary classifier for different sensory inputs, then fuse the decisions of each classifier to perform object classification.  Our proposed architecture shares similarities with the model introduced in~\cite{wang2015mmss}; however, in our setting, we did not separately train the different sensory data. Instead, the components of our architecture (i.e., CNN streams for images and MLP for depth data) were trained together.  What is more, to show the proposed architecture can achieve high accuracy, we benchmarked our results with the decision level fusion method by using a similar technique as introduced in~\cite{oh2017object}.

We present the contributions of our study in the following way. First, we propose and implement a multisensory machine learning architecture that fuses intermediate representations to create in-class shared representations and combine them with depth data to perform object recognition.   To this end, we show that this architecture improves recognition accuracy in which employing depth data performs poorly. Second,  the results show that the proposed architecture performs better to utilize the multimodal information compared to state-of-the-art fusion technique (that is, decision-level fusion) and the models that trained with single sensory data.  Third, to test further the decision level fusion and the proposed architecture, we used randomly generated images instead of the iCub's left camera inputs. In that, the results obtained in this setting shows that the decision level fusion method significantly decreased --i.e., the recognition rate lowered by $4.5 \%$. However, this setting does not affect our proposed architecture's performance as in decision level fusion; the accuracy rate only reduced by $0.11 \%$. Note that the accuracy metric was derived as dividing the total number of correct predictions (i.e.,  correctly recognized images) by the number of total predictions.

In addition to the contributions listed above, we presented a novel multisensory dataset constructed by employing the iCub robot equipped with multiple vision-based sensors. The whole dataset was shared in a public repository (see Section~\ref{reproducibility}). In doing so, we aimed to enable other researchers to develop their machine learning models, which can be used for a vision-based robotic application on the iCub robot, and compare the performance of their methods with the results reported in this paper. 

The rest of this paper is organized as follows.  We described the experiment setup, dataset specifications, and acquisition procedures in Section~\ref{exp_setup}.  Section~\ref{methods} introduces the data processing pipeline, the adopted computational models, and the proposed multisensory learning architecture for object recognition.  
The results and related discussions were presented in Section~\ref{results} and Section~\ref{discussion}, respectively. The public repository for reproducing the study and accessing the dataset was shared in Section~\ref{reproducibility}.  Finally, the conclusions and future directions of the study were outlined in Section~\ref{conclusions}.

\section{Experiment setup and Dataset description} \label{exp_setup}
\begin{figure}[ht!]
	\centering
	\begin{center}
		\includegraphics[width=0.38\textwidth, height=0.25\textheight]{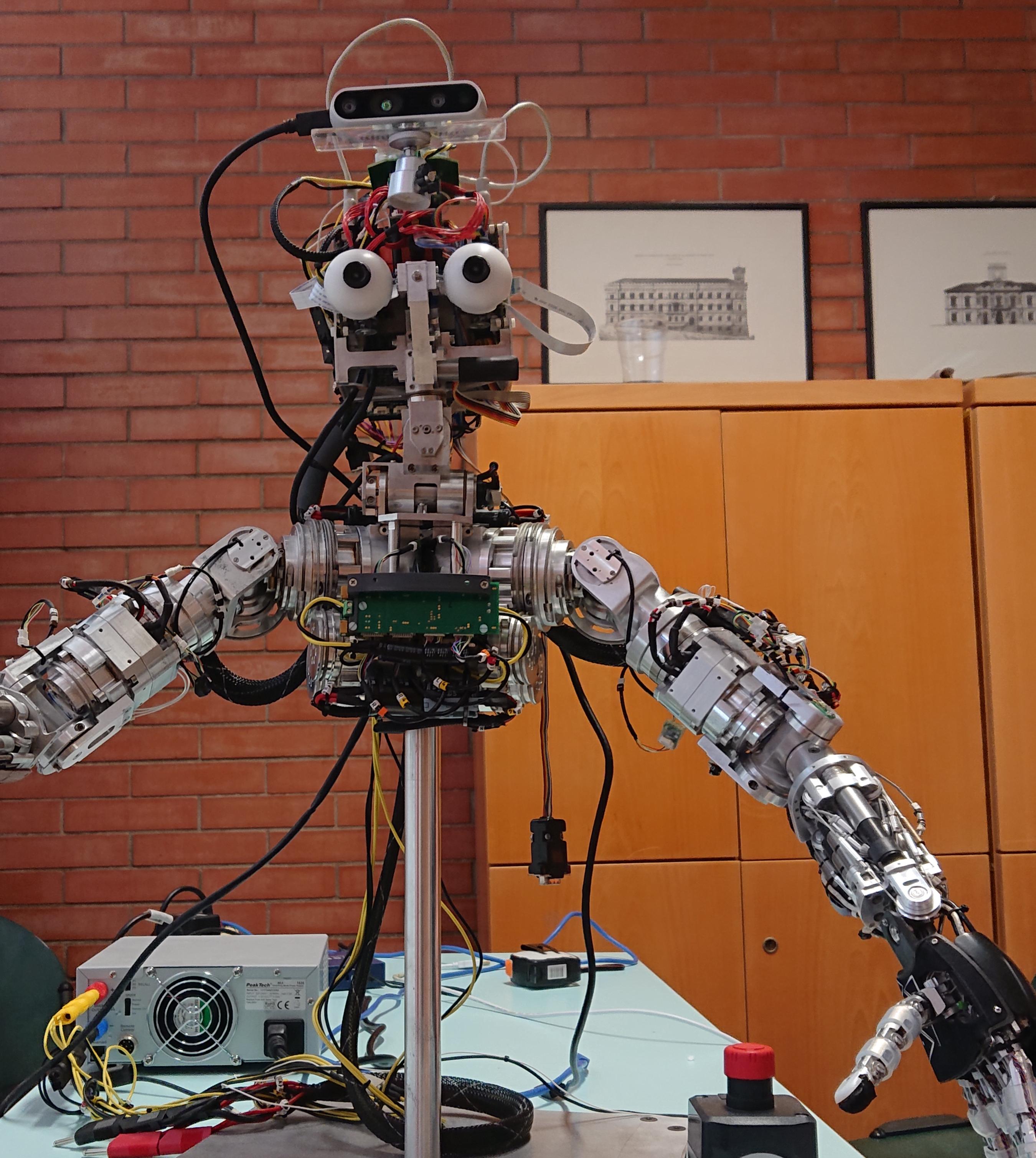}
	\end{center}
	\caption{The iCub robot equipped with three color cameras and one depth sensor.}
	\label{fig:icub}
\end{figure}

This section presents the experimental setup and the data acquisition procedure. The setup consists of a multisensory-equipped iCub robot and a motorized turntable, which was positioned in front of the robot.  As shown in Figure~\ref{fig:icub}, two Dragonfly cameras were placed on the robot's eyes, and a RealSense d435i camera was mounted on above the eyes.

\begin{figure}[ht!]
	\centering
	\begin{center}
		\includegraphics[width=0.49\textwidth]{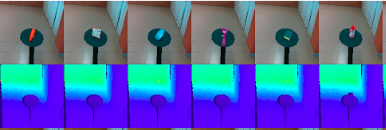}
	\end{center}
	\caption{Color images and colorized depth images.}
	\label{fig:objects}
\end{figure}

The data acquisition procedure begins with putting an object on the turntable and rotating it by approximately five degrees until completing a full rotation.  After each rotation, we recorded the color images captured from three cameras (i.e., the iCub left and right cameras, and RealSense camera) and the depth data captured from the depth sensor (i.e., depth channel of the RealSense camera). The same acquisition steps were repeated for all the objects in the dataset. Note that the data acquisition procedure is similar to those described  in~\cite{nene1996coil100},~\cite{kirtay2017dataset}.  However, in this dataset,  the number of the objects, $210$, is higher; the employed sensors used in our dataset are different from similar studies.  We depicted the color images and the colorized depth data for some of the objects in the first and second row of Figure~\ref{fig:objects}. We point out that detailed information of this dataset with hardware specifications can be found in the following technical report~\cite{kirtay2019dataset}.

At the end of the acquisition procedure,  72 different views for each object were obtained, and for each view,  72 depth and 216 color images
were collected.  In total, the whole dataset provides 60480 color and depth images. However, due to the size of the dataset, in this study, we only used the first 100 objects resulting in a total of 28800 color images and depth data matrices.  In this study, the dataset was employed to perform multimodal representation learning for object recognition. However, the models trained with this dataset can be further implemented in robotic applications: grasping type predictions for different views and learning sensorimotor schemes by mapping the images to the degree of rotation~\cite{schillaci2020intrinsic}.

\section{Methods} \label{methods}
We, here, introduce the methods for data processing and object recognition.  In the first subsection, we describe the steps to preprocess color images and depth data. In the second subsection, we present the recognition algorithms that use preprocessed color and depth data as inputs. To benchmark the performance of the recognition algorithms, we first separately employed convolutional neural networks (CNNs) for color data and a multilayer perceptron (MLP) for depth data. Then, we provide the results for multimodal learning methods: decision level fusion as a state-of-the-art fusion technique and intermediate representation fusion, which is our proposed method. 

We note that the reasons why we use the MLP for depth modality are two-fold. First, instead of implementing hand-engineered preprocessing methods such as colorizing depth data and deriving the object mask by combining pixel values, we leverage the actual (raw) distance information.
To this end, we test the performance of implemented multimodal methods without relying on intensive preprocessing procedures. However, we will compare the results presented in this paper with different preprocessing methods for color and depth data for future studies.  Second, we assess multisensory learning architecture's performance in a setting in which using depth-only data leads to poor performance for object recognition.

\subsection{Preprocessing color and depth data}
\begin{figure}[ht!]
	\centering
	\begin{center}
		\subfigure[Color image]{\label{confmats:a}\includegraphics[width=0.235\textwidth]{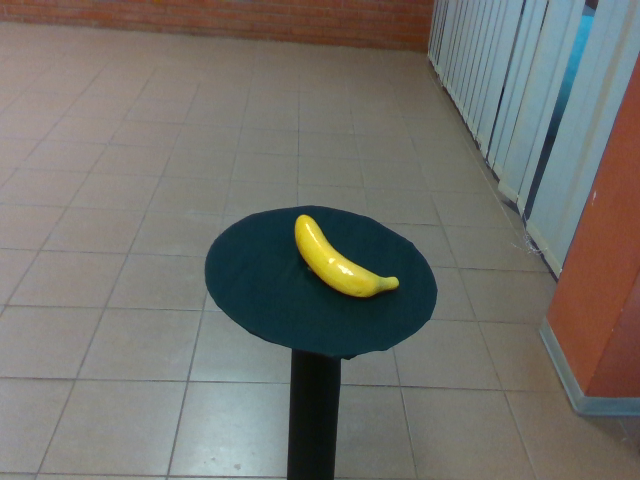}}
		\subfigure[Grayscaled image]{\label{confmats:b}\includegraphics[width=0.235\textwidth]{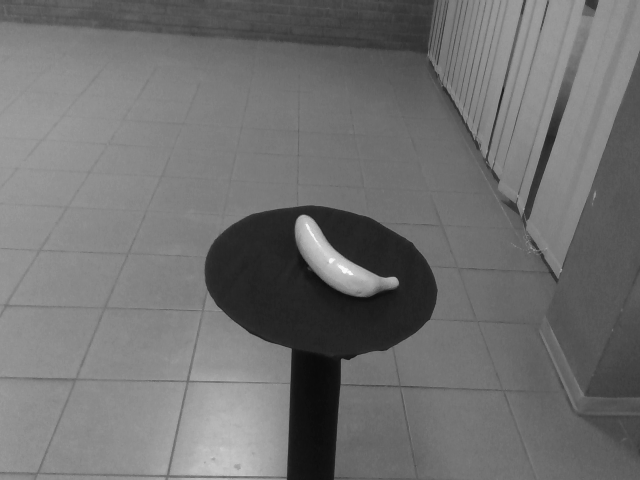}}

	\end{center}  
	\caption{Color and grayscale images of the same object.}
	\label{fig:cg_images}
\end{figure}

We performed separate preprocessing steps for the color and depth images. The color (RGB) images (whose original resolution was  $640 \times 480$ pixels) were first grayscaled to obtain a single-channel matrix of the images and then downsized to $32 \times 32$ matrix.  Since the employed dataset was constructed in an indoor environment (that is, the background of the images are the same for all objects), we deem that the information to identify the object preserved after the grayscaling operation. As can be seen in Figure~\ref{fig:cg_images}, the object can be identified in color and grayscale versions of the image. 

Since the depth sensor provides distance (in millimeter-scale) information for a pixel value in color matrices,  the preprocessing steps for depth information differ from the color images. To downsize the depth data from a matrix with a size of $640 \times 480$ to a $32 \times 32$, we applied the bilinear interpolation method.  Then the downsized matrix flattened into a vector with a size of $ 1 \times 1024$. The input vectors (i.e., the results of the two preprocessing steps) were thus associated with an output consisting of the corresponding object identifier, which was labeled during the dataset recording phase.

Here, we emphasize that flattening the depth data without further preprocessing (e.g., colorization) might lead to losing spatial information and poor performance for object recognition. However, we aimed to show that the proposed architecture will mitigate the drawbacks of preprocessing while combining color and depth information for object recognition. 

At the end of preprocessing, we first normalized input matrices and vectors. Then, to presented cross-validated results, we randomly grouped the inputs with corresponding object ids as outputs into training, validation, and testing sets by using $50\%$, $25\%$, and $25 \%$ of the whole dataset, respectively.

\subsection{Implementation of CNNs and MLP}
In this section, we introduced the implementation details of the convolutional neural networks for color modality and multilayer perceptron network for depth modality. We note that the source code, data flow diagrams, and model parameters of the networks were shared in a public repository (see Section~\ref{reproducibility}).

\subsubsection{Convolutional neural networks for preprocessed color images} \label{cnn}
In order to perform object recognition using color images for each camera, we employed a stream convolutional neural network (CNN) to extract non-hand-crafted representations (i.e., features).  The CNN stream consists of three consecutive convolutional layers -- where the first layer and the remaining layers consist of 32 and 64 filters, respectively-- in which the ReLU activation function applied after convolution operation, then $2 \times 2$ max-pooling operation implemented for downscaling. After these operations, the outputs were flattened into a single vector to be fed into a fully connected layer that has 128 hidden units with the ReLU activation function.

The whole CNN was trained for 600 epochs to minimize the categorical cross-entropy cost function via the Adam optimization algorithm~\cite{kingma2014adam}, and an early stopping condition was added to monitor validation loss.  In that, if the validation loss does not improve for a certain threshold, $0.01$, for 20 epochs, then we terminate the training. We emphasize that the recognition results were obtained from the test set.  

We note that the explained CNN stream was applied for grayscaled color images that were obtained from the iCub's left and right cameras and the mounted RealSense d435i color camera.

\subsubsection{Multilayer perceptron network for depth data} \label{mlp} 

To recognize objects by using depth data, we formed a perceptron network with three hidden layers in which each layer consists of 256 units with Rectified Linear Unit (ReLU) as the activation function. The network was trained for 600 epochs to minimize cross-entropy cost function by updating the weights via the Adam optimization algorithm. Note that we use the same early stopping procedure for 150 epochs with CNNs, and the recognition results were reported by employing test set data.

The reason why we implemented a multilayer perceptron algorithm to process depth data instead of a convolutional neural network is twofold. On the one hand, we avoid computationally intensive preprocessing (e.g., colorization of the depth data). On the other hand, we test our proposed architecture in a simple setting that does not require hand-crafted processing features such as colorizing depth images and extracting depth-mask of the object.

\subsection{Multisensory learning methods for object recognition}
In this subsection, we describe the implementation of multimodal data fusion techniques in the following way. First, we introduce a state-of-art fusion technique -- namely, decision (or late) level fusion. This method was applied to train models that accept sensory input separately, then fusing the decision made by each model to form a final decision. Second, we describe our implementation procedure of the proposed architecture -- i.e., intermediate representation fusion. Unlike decision level fusion, our method extracts features from different modalities then combine them to perform object recognition. By doing so, we aimed at benchmarking recognition results with different models and assessing our proposed architecture performance.

\subsubsection{Decision level fusion for multisensory object recognition} 
To benchmark the results with a state-of-the-art multimodal fusion method, we employed the decision (or late) level fusion.  This fusion technique is applied to exploit the performance of the models (i.e., CNNs and MLP) while mitigating the effect of the poorly performed learner. Here, decision level fusion refers to the collection of decisions from three streams of the CNNs in which each CNN was trained by employing different datasets for the same modality.
\begin{figure}[ht!]
	\centering
	\begin{center}
		\includegraphics[width=0.49\textwidth]{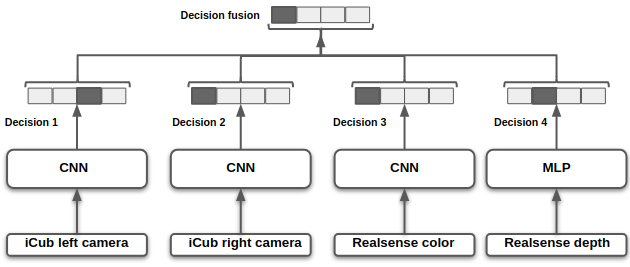}
	\end{center}
	\caption{Multisensory decision level fusion. Decision vector lengths were visualized for 4 objects, but in our experiment, each decision vector has a size of $1\times100$ where each element represents a probability values over the object ids. The values of the output vector were obtained by performing cross-entropy loss. }
	\label{fig:decision}
\end{figure} 

Figure~\ref{fig:decision} illustrates the data flow to construct decision fusion by employing the decisions made by different models. In this architecture,  a decision vector refers to a discrete probability distribution of objects. The object which has the highest probability is considered as a recognized object. In our setting, we applied CNNs for the color and MLP for the depth data to obtain decision probabilities from each model. 

After this step, the final decision for an object in the test set was formed by summing the decision probabilities to extract the recognized object id, which has the highest value. Note that normalizing decision probabilities will also yield the same results. Since we used the decisions that were generated by the models introduced in Section~\ref{cnn} and~\ref{mlp}, the training procedure for each model is the same as described in these subsections. Concretely, each model was trained separately, and then the decision vectors were fused by using the set objects.

\subsubsection{Intermediate fusion for multisensory object recognition}
To explore further multimodal methods, we structured the proposed architecture to derive two different representations: in-class shared representation and joint representation. By in-class shared representations, we refer to the representations formed by merging the last layers of convolutional neural networks that accept color data as inputs.  We extract joint representations by combining the outputs of three streams CNNs with the last layer activation of the MLP, which consists of three hidden layers. 

\begin{figure}[ht!]
	\centering
	\begin{center}
		\includegraphics[width=0.49\textwidth]{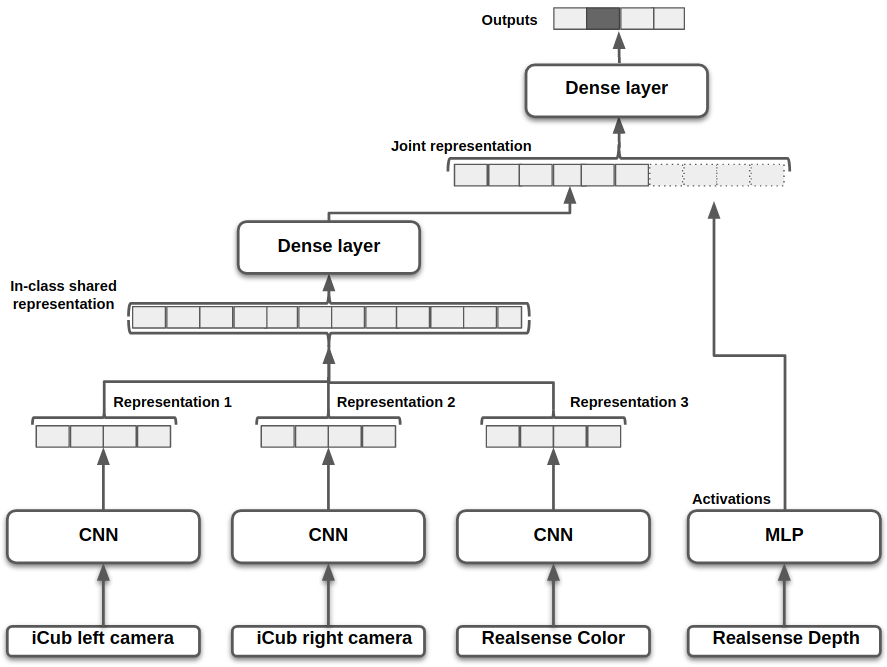}
	\end{center}
	\caption{The proposed intermediate representation fusion architecture. The output vector lengths were visualized for 4 objects,  but in our experiment, each decision vector has a size of $1\times100$ where each element represents a probability values over the object ids. The values of the output vector were obtained by performing cross-entropy loss.}
	\label{fig:intermediate}
\end{figure} 
To be concrete, as shown in Figure~\ref{fig:intermediate}, to form joint representation, we concatenated the last layer of the multilayer perceptron network with the shared representation formed by three CNN streams.  We transfer the joint representation to a densely connected layer to learn the object's representations from different sensors: color and depth.  

As the last step, we provided the joint representation to the input of a fully connected layer to extract object id, which has the highest probability, to perform object recognition.  We highlight that this architecture was trained for 600 epochs with an early stopping condition, same as with single CNN implementation, to minimize the cross-entropy cost function by employing the Adam optimization algorithm to update the network's parameters.

\section{Results} \label{results}
This section presents the results obtained by employing the machine learning models that use single and multiple sensory inputs for object recognition. To evaluate the results, we first provide the numeric values for recognition performance metrics: accuracy, precision, recall, and F1 scores. The derivation of these metrics was adopted from ~\cite{scikit}. Then, we illustrate confusion matrices as heatmaps to visually describe the recognition performance of the models.

We additionally compared the performance of the proposed architecture and decision level fusion in a setting where the iCub left camera inputs converted to the matrices where their elements were randomly assigned between 0 and 255.  In this way, we aimed to show that this approach will provide a use case to evaluate the performance of multimodal learning fusion methods in which sensory information can not provide useful information for object recognition.

\begin{table}[h!]
  \centering
	\begin{tabular}{l*4c}  
      &  Accuracy & Precision & Recall & F1-score \\
    \hline \\
 
    iCub left camera (CNN)      &0.9517 & 0.9560 & 0.9517 & 0.9520 \\
	iCub right camera (CNN)     & 0.9317 & 0.9465 & 0.9317 & 0.9331 \\
	RealSense color  (CNN)     & 0.9422 & 0.9485 & 0.9422 & 0.9427  \\
	RealSense depth  (MLP)     & 0.4256  & 0.5511 & 0.4256 & 0.4425 \\
	Decision level fusion & 0.9689  & 0.9710 & 0.9689 & 0.9690\\
	Intermediate fusion   & 0.9822  & 0.9842  & 0.9822 & 0.9825 \\	
	 
    \hline
  \end{tabular}
  \caption{Average values of recognition metrics based on sensor type and fusion architecture.}\label{table:metrics}
\end{table}
Table~\ref{table:metrics} shows the weighted average values of the recognition metrics for each corresponding model grouped by the sensor name and employed methods for object recognition.  Here, the number of correct recognitions in the test set used as support values (i.e., weights).  
The first column of Table~\ref{table:metrics}  presents the sensor names for single sensory input and multimodal data fusion techniques.  In particular, the row titled ``Intermediate fusion" row indicates our proposed architecture. The rest of the columns show the values for accuracy, precision, recall, and F1 score, respectively.  Although we evaluate the results based on accuracy and F1 scores, the other metrics were provided for further benchmarking purposes for our future studies. We note that the accuracy values derived as the number of the correctly recognized objects in the test set divided by the number of all prediction and the F1 score calculated as the harmonic mean of precision and recall values to explain the effect of two metrics in a single value.

Based on Table~\ref{table:metrics} entries, using preprocessed color data as inputs of convolutional neural networks achieves close performance to each other in terms of accuracy, albeit these inputs were obtained from a different camera. The accuracy ranging from $93 \%$ to $95 \%$. The same trend can also be observed for the F1 score.  Employing the depth data as input of a multilayer perceptron method yields $42 \%$ accuracy, which is lower than the color data, but it is substantially higher than chance level $0.01$.  We emphasize that this accuracy rate can be further improved by applying preprocessing techniques: extracting the region of interest and colorizing the depth data.  We explain the methods can be performed to improve the recognition rate by using depth modality in Section~\ref{discussion}.  We, here,  conclude that object recognition can be achieved in a single sensory setting by using color modality with higher accuracy than depth data.

Although high recognition accuracy can not be achieved using only depth modality, we exploit the effect of using depth data in multimodal learning methods.  The last two rows of Table~\ref{table:metrics} obtained by applying the decision level fusion and intermediate representation fusion, which is presented as the last row of the table. From these entries, we conclude that our proposed multisensory learning architecture provides the highest accuracy, and the difference between decision level fusion, as the closest competitor, is $1.3 \%$.  The accuracy difference between intermediate representation fusion and the MLP method that uses depth data as input is $ 55 \%$. Overall, multimodal learning methods achieve a high recognition rate, albeit including depth data, which led to poor performance for the MLP method.

To further analyze the performance of the proposed multimodal architecture in a different setting, we replaced the iCub's left camera's images with randomly generated (grayscale) data by using a uniform distribution. After manually constructing inputs for training, validation, and test sets, we used them as inputs of decision level fusion and our proposed architecture for recognizing objects. We followed the same procedures training, cross-validation, and testing described in Section~\ref{methods}.

\begin{table}[h!]
  \centering
	\begin{tabular}{l*4c}  
      &  Accuracy & Precision & Recall & F1-score \\
    \hline \\
 
	Decision level fusion & 0.9233  & 0.9419 & 0.9233 & 0.9262 \\
	Intermediate fusion   & 0.9811  & 0.9826  & 0.9811 & 0.9813 \\	
	 
    \hline
  \end{tabular}
  \caption{Multimodal fusion methods perfomance metrics which were obtained by employing random values instead of the iCub left camera's color images.}\label{table:metrics1}
\end{table}
Table~\ref{table:metrics1} shows the recognition metrics obtained by employing decision level fusion and intermediate fusion methods. 
The first column of Table~\ref{table:metrics1} indicates the performed methods, and the remaining columns show the derived values of accuracy, precision, recall, and F1 score, respectively. As can be seen in the corresponding rows of  Table~\ref{table:metrics} and Table~\ref{table:metrics1}, using randomly generated inputs instead of color images, leads to a decreasing recognition rate by $4.5 \%$. 
However, we report that for our proposed architecture, the recognition accuracy does not significantly change, only $0,11 \%$,  by using iCub's left camera input and employing random data. These results indicate that our intermediate fusion approach extracts robust features to mitigate the effect of random noise in one
\begin{figure*}[ht!]
	\centering
	\begin{center}
		\subfigure[The iCub left camera]{\label{confmats:a}\includegraphics[width=0.33\textwidth]{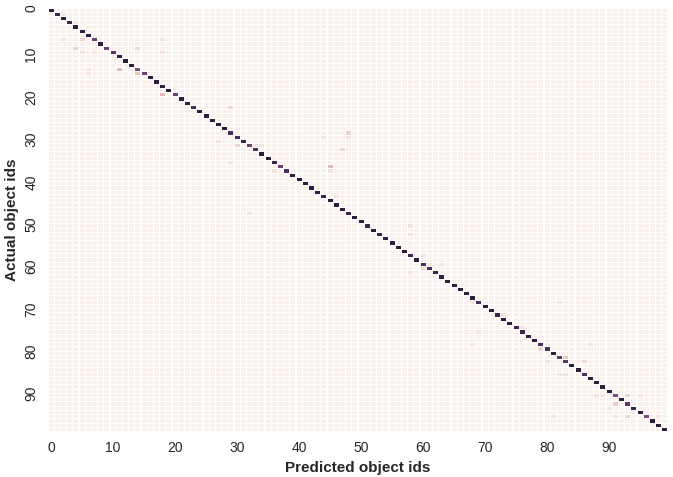}}
		\subfigure[The iCub right camera]{\label{confmats:b}\includegraphics[width=0.33\textwidth]{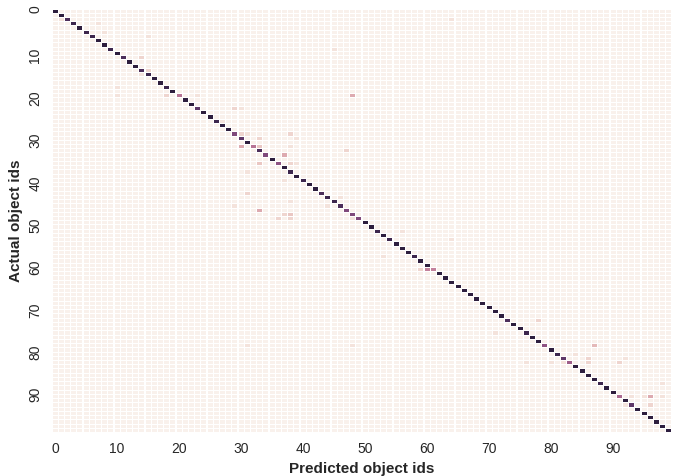}}
		\subfigure[RealSense color]{\label{confmats:c}\includegraphics[width=0.33\textwidth]{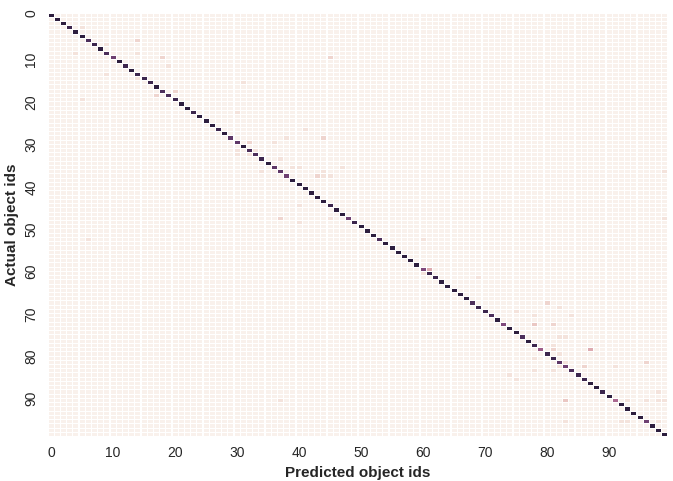}} \\
		\subfigure[RealSense depth]{\label{confmats:d}\includegraphics[width=0.33\textwidth]{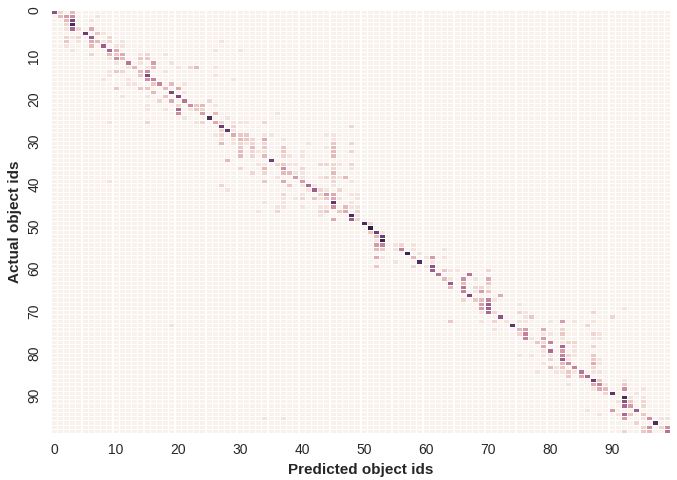}}
\subfigure[Decision level fusion]{\label{confmats:e}\includegraphics[width=0.33\textwidth]{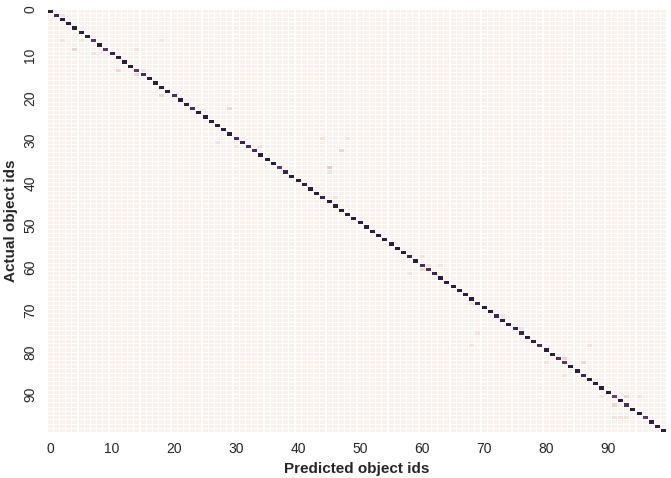}}
\subfigure[Intermediate representation fusion]{\label{confmats:f}\includegraphics[width=0.33\textwidth]{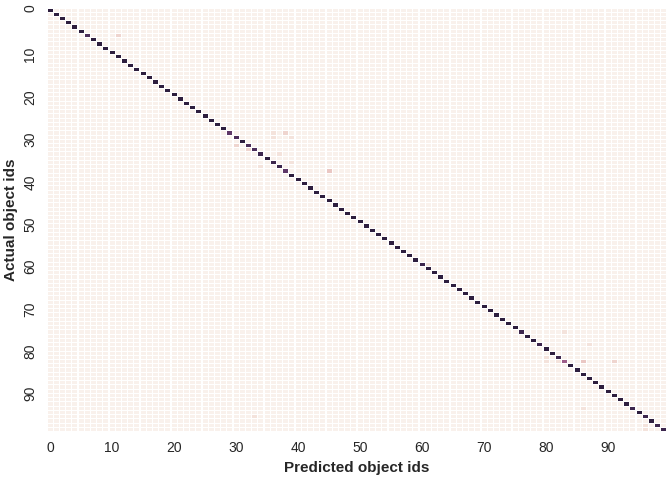}}

	\end{center}  
	\caption{Confusion matrices with unimodal and multimodal methods. The rows in these subfigures indicate that the actual object ids and the columns show the predicted object ids.}
	\label{fig:confmats}
\end{figure*}
of the sensors (i.e., the iCub's left camera) while processing multimodal data.

We hold that this setting can be seen as a realistic scenario for object recognition in robotic applications in which a sensor can be improperly calibrated or generate noisy inputs. Here, we show that the proposed architecture can overcome this deficiency in one of the sensors while performing object recognition.

To visually present the recognition results, we depicted confusion matrices as heatmaps in Figure~\ref{fig:confmats} by using the only test set data to recognize objects. The rows and columns of these matrices indicate the actual object id and recognized (or predicted) object id, respectively. The uniformity of the main diagonal of these matrices shows that the model's better performance, i.e., darker the color of the main diagonal, presents better performance. In contrast, the sparsely populated confusion matrix leads to a lower recognition rate.

The confusion matrices for color and depth inputs are shown in Figure~\ref{confmats:a},~\ref{confmats:b}, and~\ref{confmats:c}, and~\ref{confmats:d}. Figure~\ref{confmats:e} and~\ref{confmats:f} depict the confusion matrices of multimodal methods, which used both preprocessed color and depth inputs.  Based on these heatmaps,  the confusion matrix for depth modality shows sparse characteristics along the main diagonal, which corresponds to poor recognition performance. 
 For the color modality, these matrices have densely populated along the main diagonal, which presents better recognition performance than the depth modality.  The figures for the decision level fusion and intermediate representation fusion have more uniform main diagonals.  That is why these methods yield high recognition accuracy, as shown in Table~\ref{table:metrics}.  We emphasize that the characteristics of the main diagonals are in accord with the numeric values for accuracy metrics in Table~\ref{table:metrics}. For instance, the confusion matrix of intermediate representation fusion shows almost no sparsity, whereas the confusion matrix of the depth modality shows high sparse characteristics.

We highlight the following conclusions based on the entries in Table~\ref{table:metrics}, Table~\ref{table:metrics1} and the illustrated confusion matrices in Figure~\ref{fig:confmats}.  Since the characteristics of the main diagonals for the color sensors are different -- i.e., the color code of the main diagonal varies in distinct locations of the line-- each CNN model can recognize different objects well.   To exploit this observation, our intermediate representation fusion method combines the multisensor inputs in a way that captures the object's features and leads to better recognition performance.  Based on the accuracy metric in Table~\ref{table:metrics},  we realize that our proposed multimodal architecture leverages this observation better than the decision level fusion method and other benchmarked models that use single sensory input. 

What is more, the recognition metrics in Table~\ref{table:metrics1}
shows that our method's performance lowered less than the decision level fusion method in a setting where a sensory reading manually contaminated inputs with random values.  This scenario is an indication that the proposed architecture is suitable for robotic applications (e.g., multisensory guided grasping) in an environment where external noise and hardware deficiencies are unpredictable.

\section{Discussion} \label{discussion}
In this section, we discuss the steps to improve each model's performance to guide future studies. Recall that, in this paper, we aimed to present our multisensory learning architecture and provide the baseline results by employing the models trained with single sensory and multisensory inputs. 

As presented in Section~\ref{results}, the recognition rates for the color data are higher -- that is, above $90 \%$-- than the depth data. These accuracy rates indicate that the convolutional neural network achieves to form useful representations for object recognition.  We envision that the results can be further improved while neglecting the computational aspects of the training process in the following ways. The structure of the CNNs can be modified by increasing the number of filters, performing drop-out, and adding more fully connected layers. Transfer learning can also be performed for color data by adapting pre-trained models of ResNet and VGGNet to perform object recognition in the dataset employed in our study~\cite{simonyan2014very},~\cite{he2016deep}.

The steps for reducing the shortcomings by using depth data can be listed as follows. The depth data can be colorized by using the JetColor map and deep depth colorization methods~\cite{deptcolor2018},~\cite{madai2016revisiting}.  Then, a convolutional neural network can be employed to perform object recognition. By doing so, the hand-crafted depth encoding algorithms can also be applied to process raw depth data to form feature vectors before using the multilayer perceptron algorithm. For instance, the background can be removed from the depth data and color images can be employed to extract the depth mask of the object; then, the depth mask can be used as input of MLP~\cite{lai2011large}.

The multimodal learning techniques performed in this study result in better performance than the models that use single sensory inputs.  The decision level fusion exploits the multimodal data by summing the discrete probability distribution values over classes obtained from the models which use single sensory input.  Unlike decision level fusion, our proposed method achieves high accuracy by extracting robust features from the color images via CNNs and training these representations with an MLP network, which uses depth data.

Here, we conclude that the suggested methods to improve the results obtained using color and depth modalities will also enhance the performance of multimodal learning techniques.

\section{Reproducibility of the study} \label{reproducibility}
To reproduce the presented results and provide the related data -- including scripts, datasets, trained models, parameters, model diagrams, and preprocessed input data-- to the interested researchers, we used a public repository\footnote{www.github.com/msa-arxiv2020}.

\section{Conclusions} \label{conclusions}
In this study, we proposed a multisensory learning architecture, which utilizes convolutional neural networks and multi-layer perceptron method, for object recognition.  We presented the results for object recognition by employing machine learning models trained with single sensory inputs and multimodal fusion method: decision level fusion. The results indicate that our proposed architecture enhances accuracy rates compared with the benchmarked object recognition methods.  The presented results can also be considered as a baseline for other researchers who can leverage the dataset to develop multimodal machine learning models to address the object recognition task on the iCub robot.

We envision that this study can be extended in the following ways. 
First, the scaling characteristics of the proposed architecture can be assessed by adding the remaining 110 objects to the employed dataset. Second,  explainable artificial intelligence (XAI) methods can be applied in a multimodal way to interpret the role of each sensory information and contribution of each layer of the models for object recognition~\cite{samek2017evaluating}. Lastly, since the color images and depth data are associated with rotation angle, the dataset can also be employed to learn multimodal sensorimotor schemes by sensory readings (e.g., images) to the degree of rotation~\cite{schillaci2020intrinsic}.

\begin{acks}

Murat Kirtay and Verena V. Hafner have received funding from the Deutsche Forschungsgemeinschaft (DFG, German Research Foundation) under Germany's Excellence Strategy - EXC 2002/1 ``Science of Intelligence" - project number 390523135.
Guido Schillaci has received funding from the European Union's Horizon 2020 research and innovation programme under the Marie Sklodowska-Curie grant agreement No. 838861 (Predictive Robots).
\end{acks}

\bibliographystyle{ACM-Reference-Format}
\bibliography{msl_architecture}

\end{document}